# Hybrid Machine Learning Framework for Herd-Level Cattle Growth Pattern and Weight Gain Forecasting in Grazing-Based Production Systems

Muhammad Riaz Hasib Hossain [a, 1], Rafiqul Islam [b], Shawn R. McGrath [c], Md Zahidul Islam [d, e], and David W. Lamb [f, g]

[a] *School of Computing, Mathematics and Engineering, Charles Sturt University, Wagga Wagga, NSW 2650, Australia*
[b] *School of Computing, Mathematics and Engineering, Charles Sturt University, Albury, NSW 2640, Australia*
[c] *Gulbali Institute for Agriculture, Water and Environment, Charles Sturt University, Wagga Wagga, NSW 2678, Australia*
[d] *School of Computing, Mathematics and Engineering, Charles Sturt University, Australia, Panorama Avenue, Bathurst NSW 2795, Australia*
[e] *AI and Cyber Futures Centre, Charles Sturt University, Australia, Panorama Avenue, Bathurst NSW 2795, Australia*
[f] *Precision Agriculture Research Group, University of New England, NSW, Australia 2351*
[g] *Food Agility CRC Ltd, Sydney, NSW, Australia 2000.*

**Abstract**

Commercial grazing systems yield irregular livestock observations, which challenge cattle growth forecasting. The study developed a hybrid machine learning framework for herd-level cattle weight forecasting using automated sensing observations collected between 2022 and 2024 in southeastern Australia. Weekly live-weight observations, demographic variables, and lagged environmental predictors were integrated into structured forecasting datasets. Herd-level forecasting trajectories were generated through temporal aggregation of animal-level predictions. Four hybrid architecture families were evaluated, including residual, stacked, cascade, and ensemble-assisted frameworks. ARIMA, LSTM, and GRU models were used as comparative baselines. Independent testing demonstrated strong predictive agreement across multiple forecasting horizons. The cascade GB→RF→NN architecture achieved the best performance, with a test $R^2$ of 0.889, RMSE of 21.319 kg, and MAE of 15.462 kg. Hybrid architectures maintained stronger robustness than recurrent sequential models under sparse observation conditions. Forecasting error increased progressively across extended prediction horizons. Feature-importance analysis identified animal age, rainfall, and temperature as dominant predictors influencing herd-level growth forecasting. The proposed framework may support feed allocation, grazing management, and livestock marketing decisions under heterogeneous sensing environments.

## 1. Introduction

Grazing-based beef production systems contribute substantially to food security, rural economies, and agricultural sustainability (Thornton & Herrero, 2014). Herd-level cattle growth represents an important production indicator because growth performance influences pasture utilisation, feed allocation, stocking strategy, and livestock marketing decisions (Tedeschi et al., 2021). Commercial grazing enterprises increasingly require predictive decision-support systems that can respond to climatic variability and fluctuating animal growth performance (Herrero et al., 2017).

Conventional live-weight monitoring in grazing enterprises commonly relies on infrequent yard weighing or visual assessment (Hossain et al., 2025b). Such approaches increase labour demand and animal stress and provide limited temporal resolution (Dickinson et al., 2013). Automated step-on-off (SOO) weighing systems support low-intervention livestock monitoring under commercial paddock conditions (Aquilani et al., 2022).

[1] Corresponding author: Muhammad Riaz Hasib Hossain (email: muhossain@csu.edu.au)

Higher observation frequency improves the assessment of herd growth dynamics under changing pasture and climatic conditions (Parsons et al., 2023).

Cattle growth within grazing systems is influenced by interacting biological and environmental processes. Animal age, forage availability, rainfall variability, and temperature collectively influence herd growth trajectories (Owens et al., 1993). Climatic variability affects pasture productivity and animal intake, which subsequently influence live-weight gain (Hasan et al., 2025; Hossain et al., 2025a). Herd-level growth trajectories may provide indicators of pasture performance, nutritional adequacy, and production efficiency within grazing enterprises (Tedeschi et al., 2004).

Most livestock forecasting studies apply sequential forecasting models using regularly sampled observations collected under controlled production systems (Ruchay et al., 2022). Commercial grazing systems yield irregular observations because cattle voluntarily interact with weighing platforms (Hasan et al., 2025). Uneven temporal spacing and inconsistent visitation frequency introduce behavioural variability into animal-level measurements (Aquilani et al., 2022). Herd-level aggregation may reduce stochastic variation and improve operational interpretation across grazing intervals (Tedeschi et al., 2021).

Commercial grazing systems increasingly generate heterogeneous livestock-sensing datasets through automated monitoring platforms integrated into precision livestock farming systems (García et al., 2020). Recent forecasting studies mainly examine regular temporal sequences collected under controlled production environments (Ruchay et al., 2022). Limited research has evaluated biologically informed forecasting under irregular grazing conditions containing inconsistent observation intervals and incomplete temporal continuity (García et al., 2020). Hybrid machine learning (ML) approaches may improve forecasting stability by integrating nonlinear biological and environmental relationships under sparse sensing conditions (Morota et al., 2018). Operational herd-level forecasting may also support grazing management decisions by improving the representation of enterprise-scale production dynamics (Tedeschi et al., 2025). Existing recurrent forecasting architectures often require stable temporal sequences and high observation density (Che et al., 2018). Such assumptions remain difficult to maintain in the context of voluntary livestock visitation behaviour associated with automated SOO weighing systems. Forecasting instability under sparse sensing conditions remains under-examined in commercial grazing environments (García et al., 2020).

Operational livestock datasets contain irregular visitation intervals and incomplete temporal sequences (Emmanuel et al., 2021). Such characteristics complicate forecasting under commercial grazing conditions (Che et al., 2018). Existing livestock forecasting studies mainly examine controlled production systems with stable sampling intervals (Hossain et al., 2025b). Limited studies evaluate forecasting robustness under irregular grazing environments (García et al., 2020). Table 1 summarises the irregular observational characteristics calculated from animal-specific visitation intervals across the livestock sensing dataset.

**Table 1**
Irregular sensing characteristics of commercial SOO livestock datasets collected under commercial grazing conditions.

| Metric | Value |
|---|---|
| Mean observation interval | 3.8 days |
| Median weekly observations per animal | 2.9 observations |
| Missing weekly intervals | 18.6% |
| Observation interval coefficient of variation | 42.3% |
| Sparse observation periods (>7 days) | 14.2% |
| Observation density reduction during dry periods | 21.5% |

Note: Observation irregularity metrics were calculated using animal-specific visitation intervals derived from temporally filtered livestock observations collected between 2022 and 2024. The observation interval coefficient of variation (CV) quantified variability across successive animal-level weighing intervals within the monitored population. Sparse observation periods represented intervals containing fewer than two valid weekly observations per animal following quality-control procedures. Observation density reduction was quantified during periods of

reduced pasture availability and elevated climatic variability. Reduced cattle visitation frequency during dry periods was observed under the voluntary SOO weighing system operating within the commercial grazing environment.

Biologically informed tabular transformation may improve livestock forecasting under irregular observation conditions (Liakos et al., 2018). Lagged environmental predictors may represent delayed pasture and climatic responses without requiring continuous temporal sequences (Pearson et al., 2021). Hybrid ML architectures may improve predictive stability by integrating complementary modelling mechanisms that can represent nonlinear biological interactions (Kamilaris & Prenafeta-Boldú, 2018). Weekly forecasting intervals provide a practical temporal resolution for grazing management planning (Aquilani et al., 2022).

The study evaluated residual, stacked, cascade, and ensemble-assisted ML architectures using multi-year livestock observations collected from a commercial grazing enterprise. Comparative evaluation examined forecasting accuracy, forecasting stability across horizons, computational efficiency, and robustness under irregular livestock observation conditions.

The objectives of this study were to: (i) develop an operational herd-level forecasting framework using structured datasets derived from irregular livestock observations; (ii) compare hybrid ML architecture families in terms of predictive accuracy and computational efficiency; (iii) evaluate forecasting robustness across multiple operational prediction horizons and observation-density conditions; and (iv) identify dominant biological and environmental predictors influencing herd-level cattle growth dynamics.

## 2. Materials and Methods

The methodology included data collection, quality control, weekly aggregation, dataset construction for forecasting, model training, and independent validation. Figure 1 presents the workflow for weekly herd-level cattle growth forecasting, which uses automated SOO observations and environmental predictors.

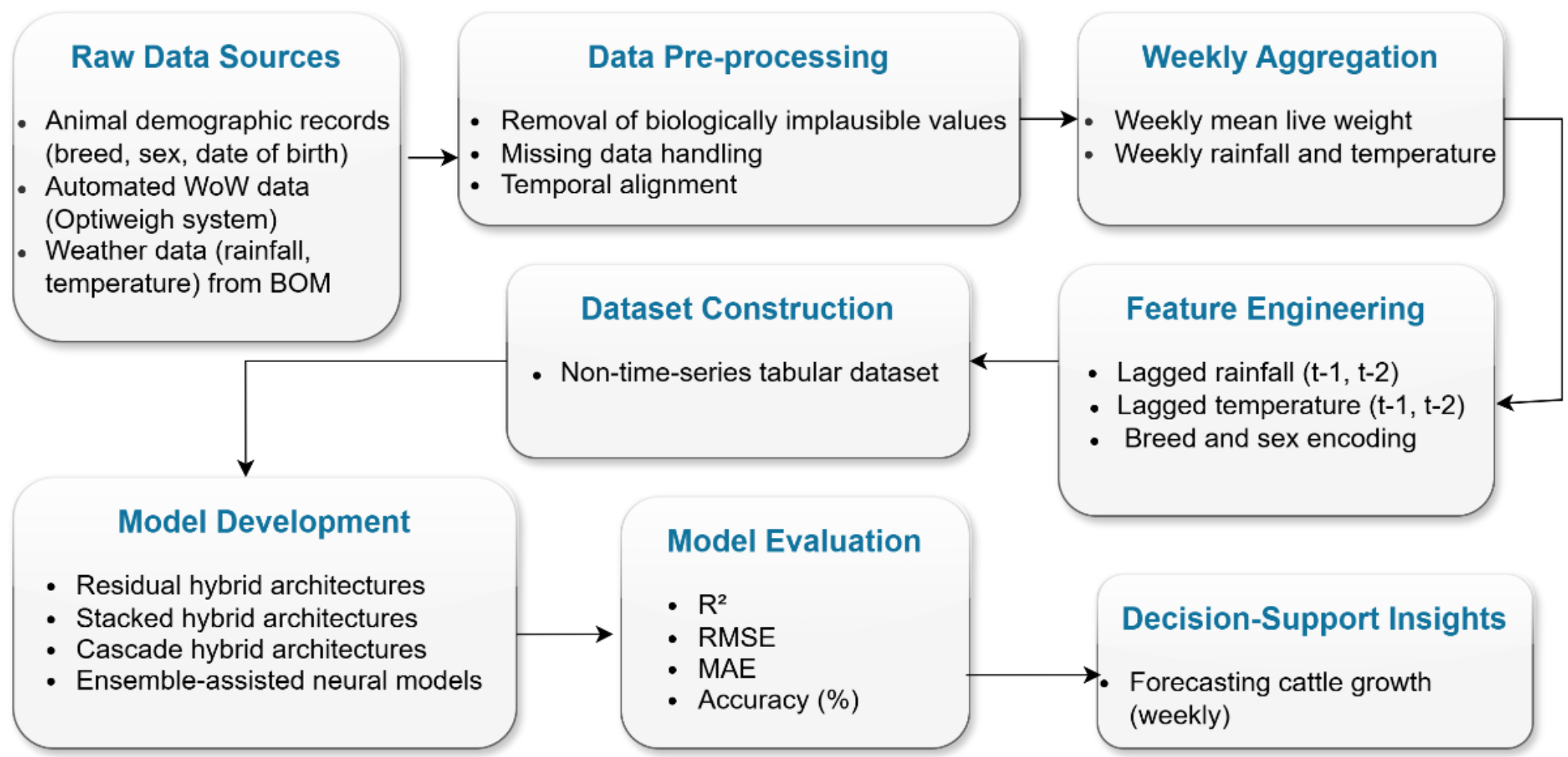

Fig. 1. Methodological workflow for weekly herd-level cattle growth forecasting using hybrid ML architectures.

### 2.1. Study site and data sources

The study was conducted at the Charles Sturt University (CSU) Farm, Wagga Wagga, Australia, between 2022 and 2024 under temperate grazing conditions typical of southeastern Australian beef systems. The site enabled the evaluation of forecasting performance under commercial grazing conditions.

Three primary data streams were integrated for model development, including animal demographic records, automated live-weight observations, and climatic measurements. Demographic records included genotype, sex, and date of birth for each monitored animal. Live-weight observations were collected using the Optiweigh automated SOO weighing system (Parsons et al., 2023). Environmental variables included daily rainfall and temperature measurements obtained from the nearest Australian Bureau of Meteorology (BOM[2]) station.

Integrated livestock and environmental observations represented biological and climatic factors influencing cattle growth dynamics. The dataset included 960 Angus and Limousin cattle monitored under commercial grazing conditions.

The CSU Animal Ethics Committee approved all data collection procedures under approval numbers A21414 and A22404. Experimental procedures complied with national guidelines for the ethical use of animals in agricultural research.

### 2.2. Data pre-processing and weekly aggregation

Raw live-weight observations were filtered to remove biologically implausible short-duration fluctuations. Consecutive fluctuations exceeding ±15 kg within 24 hours were excluded to reduce transient sensor instability and weighing artefacts (Alawneh et al., 2011). Weekly mean live weight was subsequently calculated for each animal.

Daily rainfall observations were aggregated into cumulative weekly totals. Daily temperature observations were aggregated into weekly mean values. Environmental observations were temporally aligned with livestock records to maintain consistency across forecasting intervals.

Missing values associated with short temporal gaps of seven days or less were imputed using linear interpolation. Longer temporal gaps were imputed using k-nearest-neighbour regression. Imputation procedures were applied independently within training and testing subsets to minimise information leakage.

Each animal required at least four valid live-weight observations per month to support stable weekly aggregation. A total of 686 cattle passed quality control and filtering for modelling analysis, generating 21,406 weekly observations across 2022 to 2024.

Weekly live-weight gain was calculated only for secondary biological interpretation analyses examining genotype- and sex-associated growth variability. Weekly animal-level live weight remained the primary supervised forecasting target throughout model training and evaluation.

[2] The website of BOM (http://bom.gov.au)

*2.3. Herd-level aggregation framework*

Operational grazing enterprises commonly implement decisions at the herd scale (Berckmans, 2017). Weekly animal-level predictions were aggregated into herd-level trajectories to reduce short-term behavioural variability.

Weekly herd-level live weight was calculated as follows:

$$H_t = \frac{1}{N_t} \sum_{i=1}^{N_t} W_{i,t} \quad (1)$$

where $H_t$ represents herd-level live weight at week $t$ , $W_{i,t}$ represents the weekly live weight of animal $i$, and $N_t$ represents the number of valid cattle observations recorded during week $t$.

Herd-level temporal smoothing was subsequently evaluated using rolling variance and CV analyses across weekly forecasting intervals. Variance reduction following aggregation was calculated as:

$$VR = \frac{\sigma_i^2 - \sigma_h^2}{\sigma_i^2} \times 100 \quad (2)$$

where $VR$ represents variance reduction percentage, $\sigma_i^2$ represents individual-animal variance, and $\sigma_h^2$ represents herd-level variance after temporal aggregation.

CV was additionally evaluated before and after aggregation to quantify biological stability under irregular sensing conditions. Herd-level aggregation reduced stochastic fluctuations associated with irregular weighing frequency and improved the representation of seasonal growth dynamics in grazing systems.

The herd-level aggregation framework improved compatibility between irregular livestock-sensing observations and enterprise-scale forecasting workflows under commercial grazing conditions.

*2.4. Forecasting-oriented structured representation of irregular livestock observations*

A structured forecasting framework transformed irregular livestock observations into supervised learning datasets suitable for herd-level prediction. Structured tabular representation improved compatibility with irregular temporal observations (Liakos et al., 2018). Weekly livestock observations were transformed into indexed forecasting records containing biological predictors, lagged environmental variables, historical live-weight indicators, and temporally aligned forecasting targets (García et al., 2020). Structured forecasting records supported prediction under heterogeneous visitation intervals and incomplete observations in the current study.

Each weekly livestock observation was transformed into a temporally indexed forecasting record containing biological, environmental, and historical predictors. Animal age in weeks was incorporated to represent physiological growth progression (Owens et al., 1993). Environmental predictors included rainfall and temperature variables from the current week, along with one-week (t-1) and two-week (t-2) lag intervals. Lag selection reflected delayed pasture growth and short-term forage availability associated with climatic variability in grazing systems (Rojas-Downing et al., 2017). Historical live-weight indicators were incorporated to improve short-term forecasting stability under heterogeneous observation intervals (Ruchay et al., 2022). Categorical variables, including genotype and sex, were encoded using one-hot encoding to maintain compatibility with ML forecasting algorithms.

*2.4.1. Mathematical forecasting transformation framework*

Weekly livestock observations were transformed into supervised forecasting records using temporal indexing procedures. The transformed forecasting record for animal $i$ at week $t$ was represented as follows:

$$X_{i,t} = [W_{i,t-1}, A_{i,t}, R_t, R_{t-1}, R_{t-2}, T_t, T_{t-1}, T_{t-2}, G_i, S_i] \quad (3)$$

where $W_{i,t-1}$ represents historical live weight, $A_{i,t}$ represents animal age, $R_t$ represents weekly rainfall, $R_{t-1}$ and $R_{t-2}$ represent one-week and two-week rainfall lag variables, $T_t$ represents weekly temperature, $T_{t-1}$ and $T_{t-2}$ represent lagged temperature variables, $G_i$ represents the genotype category, and $S_i$ represents the sex category. Lagged environmental variables represented delayed climatic responses associated with pasture growth and forage availability.

The forecasting target for prediction horizon $h$ was defined as:

$$Y_{i,t+h} = W_{i,t+h} \quad (4)$$

where $Y_{i,t+h}$ represents the live-weight forecasting target for animal $i$ at prediction horizon $h$, and $W_{i,t+h}$ represents the observed live weight at future week $t + h$. Forecasting horizons ranged from 1 to 24 weeks to evaluate short-term and extended forecasting performance under operational grazing conditions.

Temporal aggregation was subsequently applied to weekly prediction outputs to derive herd-level forecasting trajectories. Aggregation reduced short-term behavioural variability under commercial grazing conditions.

*2.4.2. Forecasting transformation workflow*

The forecasting transformation workflow converted irregular livestock-sensing observations into supervised learning datasets suitable for hybrid ML architectures. Figure 2 summarises the transformation framework applied to commercial SOO livestock observations.

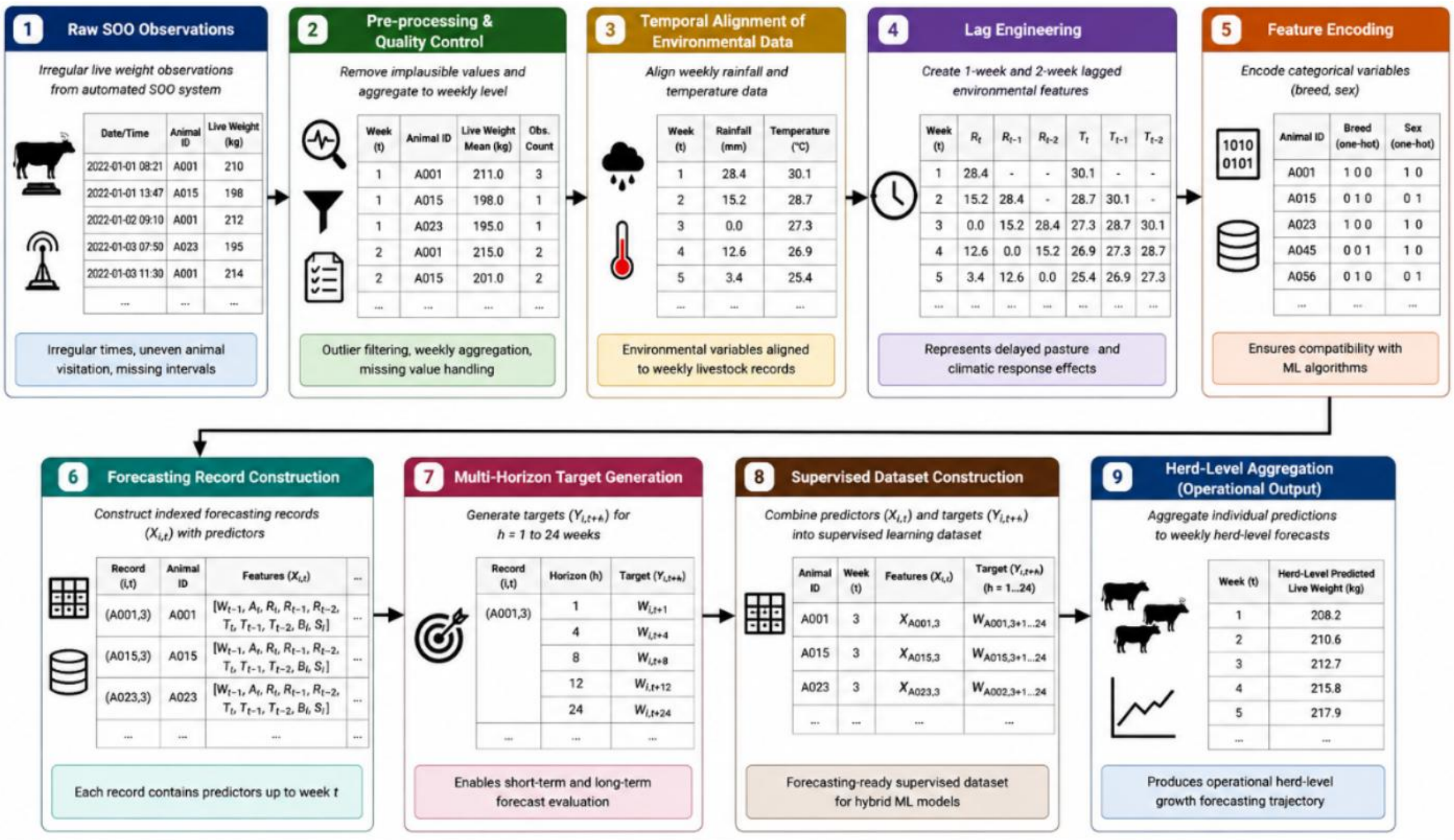

Fig. 2. Forecasting-oriented transformation workflow for irregular livestock sensing observations collected under commercial grazing conditions.

The workflow summarises pre-processing, feature engineering, supervised forecasting construction, and herd-level aggregation. Table 2 summarises the transformation workflow applied to irregular livestock observations.

**Table 2**

Forecasting-oriented transformation workflow for irregular livestock observations collected under commercial grazing conditions.

| Transformation Stage | Input Data | Transformation Procedure | Forecasting Purpose |
|---|---|---|---|
| Temporal aggregation | Raw SOO observations | Weekly averaging | Reduces short-term stochastic fluctuations |
| Quality control | Sensor observations | Outlier filtering | Removes biologically implausible measurements |
| Missing-value handling | Incomplete weekly intervals | Interpolation and KNN imputation | Maintains forecasting continuity |
| Lag engineering | Environmental variables | One-week and two-week lag generation | Corresponds to delayed pasture variation |
| Feature encoding | Genotype and sex | One-hot encoding | Supports ML compatibility |
| Forecast target generation | Weekly live weight | Temporal target indexing | Enables multi-horizon forecasting |
| Herd-level aggregation | Individual predictions | Weekly temporal averaging | Supports operational herd forecasting |

Note: The transformation workflow converted irregular livestock sensing observations into forecasting-ready supervised learning datasets suitable for operational grazing-system forecasting under heterogeneous sensing conditions.

Lagged environmental predictors reflected delayed climatic responses that affected cattle growth. Historical live-weight, biological, and environmental predictors were integrated within the forecasting dataset. Forecasting records were constructed from historical individual-animal observations collected by automated SOO systems. Weekly live weight was used as the supervised forecasting target because animal-level observations preserved sensitivity to physiological growth during training. Herd-level trajectories were derived through temporal aggregation of weekly prediction outputs. Forecasting horizons ranged from 1 to 24 weeks. Separate supervised datasets were constructed for each horizon. Animal sex was binary-encoded, and genotype variables were one-hot-encoded. Table 3 summarises the operational justification for structured forecasting representation under irregular livestock observation conditions.

**Table 3**
Operational justification for structured forecasting representation under irregular livestock observation conditions.

| Data Characteristic | Sequential Model Limitation | Transformation Strategy | Operational Benefit |
| --- | --- | --- | --- |
| Irregular measurement frequency | Recurrent forecasting models require stable temporal intervals | Weekly aggregation and structured forecasting records | Supports forecasting under opportunistic livestock visitation |
| Missing observations | Sequential architectures are sensitive to incomplete temporal sequences | Imputation and structured feature construction | Maintains forecasting continuity |
| Uneven sampling intervals | Sequential learning assumes uniform temporal spacing | Lagged biological and environmental predictors | Represents delayed physiological responses |
| Multi-source predictor integration | Sequential frameworks are less flexible for heterogeneous predictors | Integration of demographic, environmental, and live-weight variables | Supports operational grazing-system forecasting |

Note: Structured forecasting representation improved compatibility between irregular livestock observations and hybrid forecasting workflows.

### *2.5. Comparative forecasting models*

Comparative baseline models evaluated whether hybrid architectures improved forecasting performance under irregular sensing conditions. Sequential architectures were additionally tested under controlled observation-density scenarios. Baseline comparisons supported objective performance evaluation.

Three baseline models were implemented, including autoregressive integrated moving average (ARIMA), long short-term memory (LSTM), and gated recurrent unit (GRU). These approaches are widely used for biological and environmental forecasting tasks (Kamilaris & Prenafeta-Boldú, 2018).

ARIMA modelling was implemented using weekly aggregated herd-level live-weight observations. LSTM and GRU architectures used sequential input windows containing previous live-weight observations and aligned environmental predictors. Dropout regularisation and Adam optimisation were applied during neural-network (NN) training. Sequential observations containing missing values were interpolated before recurrent sequence construction.

Baseline sequential models were evaluated using the same training and test partitions used for the hybrid architectures. A performance comparison between the baseline and hybrid approaches was conducted using Coefficient of Determination ($R^2$), Root Mean Square Error (RMSE), and Mean Absolute Error (MAE) metrics derived from independent test datasets. Table 4 summarises the configuration settings for comparative baseline forecasting models. Architectures were compared using predictive accuracy and computational efficiency.

**Table 4**
Configuration settings for comparative baseline forecasting models evaluated under commercial grazing conditions.

| Baseline Model | Configuration | Purpose |
| --- | --- | --- |
| ARIMA | Order selection using Akaike Information Criterion (AIC) optimisation | Statistical time-series baseline |
| LSTM | Two hidden recurrent layers with dropout regularisation | Deep sequential forecasting baseline |
| GRU | Two hidden recurrent layers with dropout regularisation | Lightweight recurrent forecasting baseline |
| Optimiser | Adam optimisation algorithm | NN parameter optimisation |
| Input structure | Weekly sequential windows with environmental predictors | Multi-variable temporal learning |
| Evaluation metrics | $R^2$, RMSE, MAE | Independent forecasting evaluation |

Hybrid architectures were evaluated because grazing datasets contain nonlinear biological and environmental interactions (Morota et al., 2018). Residual, stacked, cascade, and ensemble-assisted frameworks were compared for predictive accuracy, computational demand, and forecasting robustness (Kamilaris & Prenafeta-Boldú, 2018). Figure 3 presents the evaluated hybrid ML architectures used for weekly herd-level cattle growth forecasting.

Residual hybrid architectures generated an initial prediction using a base learner. NNs were subsequently used to model residual prediction errors to refine final outputs. Residual learning enabled NNs to focus on unexplained variance rather than the complete prediction process. Models evaluated within this architecture included Gradient Boosting (GB), Random Forest (RF), Support Vector Regression (SVR), and Extreme Gradient Boosting (XGBoost) combined with residual NNs.

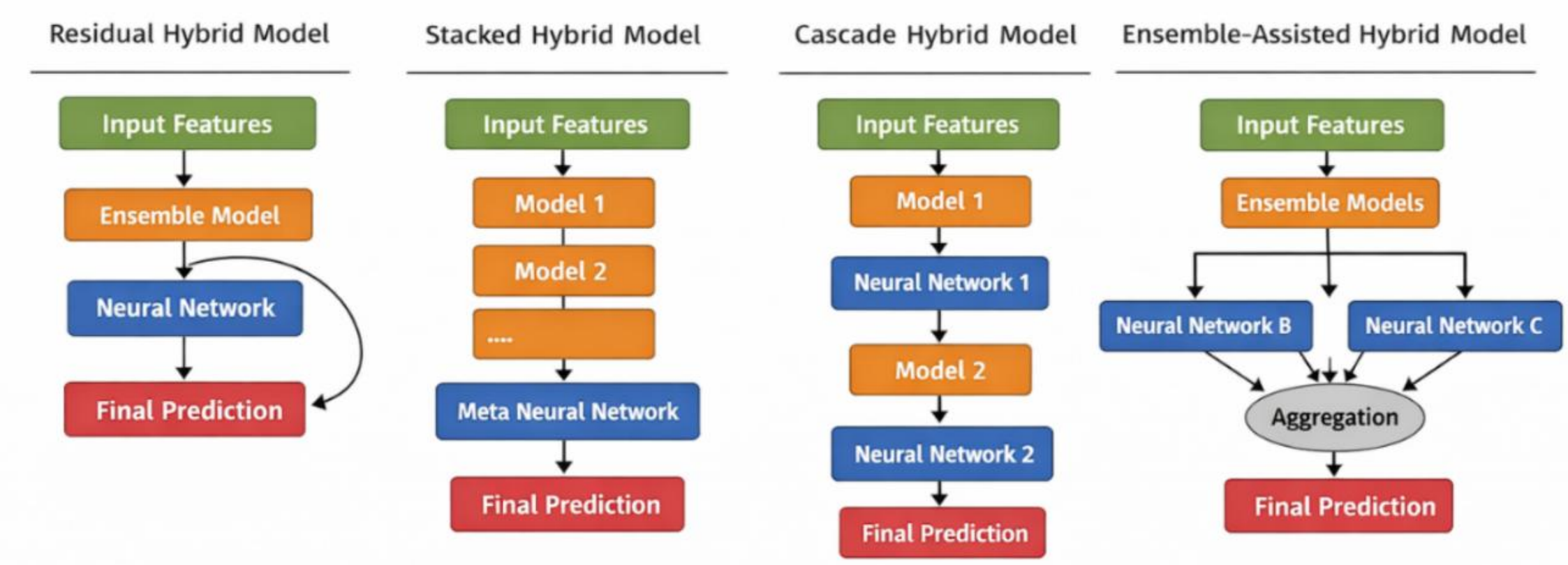

Fig. 3. Hybrid ML architecture families evaluated for weekly herd-level cattle growth forecasting.

Stacked hybrid architectures combined predictions from multiple base learners and supplied these outputs to a NN meta-learner. The structure enabled the meta-learner to exploit complementary strengths among individual models (Wolpert, 1992). Models evaluated within this architecture included combinations of RF, GB, SVR, K-Nearest Neighbour, Decision Tree (DT), and Linear Regression (LR) models integrated with NN meta-learners.

Cascade hybrid architectures sequentially propagated predictions from upstream models as additional features to downstream learners. Sequential integration enabled downstream models to refine their forecasting performance using original predictors alongside intermediate outputs. Models evaluated within this architecture included GB, RF, and SVR integrated with downstream NNs.

Ensemble-assisted neural architectures generated intermediate predictions by applying ensemble techniques to multiple base learners. Ensemble outputs were subsequently incorporated into neural networks to refine final predictions. The architecture combined the variance-reduction benefits of ensemble learning with NN representation learning to improve predictive robustness and computational flexibility in operational livestock forecasting environments (Sagi & Rokach, 2018).

## *2.6. Validation and statistical analysis*

Model evaluation employed animal-level hold-out validation with internal cross-validation applied exclusively within the training subset. Training and testing partitions were constructed using unique electronic identification (EID) codes assigned to individual cattle (Roberts et al., 2017). This approach minimised information leakage from repeated biological measurements. The final partition contained 686 unique animals, including 480 allocated to training and 206 allocated to testing. The training subset contained 14,979 weekly records, and the testing subset contained 6,427 weekly records.

Animal-level partitioning minimised information leakage from repeated biological measurements (Roberts et al., 2017). Repeated observations nevertheless remained longitudinally correlated across production intervals. Table 5 presents the characteristics and descriptive statistics of the training and testing partitions, indicating comparable data distributions between subsets. Such comparability supported robust model evaluation.

**Table 5**
Dataset characteristics and descriptive statistics for training and testing partitions.

| Variable | Training (n = 14,979 weekly records; 480 EIDs) | Testing (n = 6,427 weekly records; 206 EIDs) | Overall |
|---|---|---|---|
| Production years | 2022–2024 | 2022–2024 | 2022–2024 |
| Mean live weight (kg) | 297.90 ± 63.57 | 296.04 ± 63.89 | 297.34 ± 63.67 |
| Weight range (kg) | 150.73-530.56 | 131.25-496.08 | 131.25-530.56 |
| Mean age (weeks) | 44.10 ± 8.53 | 44.14 ± 8.48 | 44.10 ± 8.51 |
| Mean weekly rainfall (mm) | 10.82 ± 11.91 | 10.78 ± 11.87 | 10.81 ± 11.90 |
| Mean weekly temperature (°C) | 18.24 ± 4.76 | 18.24 ± 4.74 | 18.24 ± 4.75 |

Notes: Values are presented as mean ± standard deviation unless otherwise stated. Rainfall values represent cumulative weekly rainfall totals aggregated from daily meteorological observations. Partitioning was performed at the animal level using EID codes to minimise information leakage between training and testing subsets. Repeated weekly observations within individual animals nevertheless remained longitudinally correlated because cattle contributed multiple temporal observations across production intervals.

Hold-out validation was selected because the livestock dataset allowed stable estimation of generalisation performance without repeated resampling. Hyperparameters were optimised exclusively within the training subset using a grid search.

Exploratory statistical comparisons among forecasting architectures were conducted using paired t-tests on observation-level prediction errors. Pairwise comparisons evaluated whether competing architectures produced significantly different prediction errors across animal-level testing subsets. Statistical significance was assessed at $p < 0.05$, and effect sizes were quantified using Cohen's d together with 95% confidence intervals (Nakagawa & Cuthill, 2007). Residual temporal dependence from repeated longitudinal observations was accounted for when interpreting the inferential results.

Model performance was evaluated using $R^2$, RMSE, and MAE metrics. Such metrics quantified predictive performance across weekly forecasting horizons and independent testing datasets.

Higher $R^2$ values indicate stronger predictive agreement. $R^2$ was calculated as follows:

$$R^2 = 1 - \frac{\sum_{i=1}^{n}(y_i - \hat{y}_i)^2}{\sum_{i=1}^{n}(y_i - \bar{y})^2} \tag{5}$$

where $y_i$ represents observed weekly cattle weights, $\hat{y}_i$ represents predicted weekly cattle weights, and $\bar{y}$ represents the mean observed cattle weight, and $n$ represents the total number of observations.

RMSE represents the square root of the average squared difference between predicted and observed cattle weights. Lower RMSE values indicate improved predictive precision and reduced forecasting error. RMSE was calculated as follows:

$$RMSE = \sqrt{\frac{1}{n}\sum_{i=1}^{n}(y_i - \hat{y}_i)^2} \quad (6)$$

MAE measures the average absolute deviation between predicted and observed cattle weights. Lower MAE values indicate smaller average deviations between predictions and the actual values across the evaluated forecasting intervals. MAE was calculated as follows:

$$MAE = \frac{1}{n}\sum_{i=1}^{n} | y_i - \hat{y}_i | \quad (7)$$

All evaluation metrics were calculated using independent testing datasets. Identical Python-based computational environments were used across experiments. Prediction uncertainty was assessed using confidence intervals derived from training-validation procedures.

### *2.7. Herd-level stability and variance reduction analysis*

Additional statistical analysis quantified the improvement in biological stability associated with herd-level aggregation. Weekly variance, standard deviation, and CV were evaluated before and after temporal aggregation of individual-animal observations. The CV was calculated as follows:

$$CV = \frac{\sigma}{\mu} \times 100 \quad (8)$$

where $CV$ represents the coefficient of variation, $\sigma$ represents standard deviation, and $\mu$ represents mean live weight. Lower CV values indicated improved biological stability following aggregation.

Forecasting performance was additionally compared between individual-animal prediction outputs and aggregated herd-level trajectories using RMSE and MAE metrics derived from independent testing datasets.

### *2.8. Observation density and missingness sensitivity analysis*

Additional robustness experiments were conducted to evaluate forecasting stability under varying observation-density conditions commonly observed in commercial grazing systems (Hasan et al., 2025). Controlled observation sparsity scenarios were developed to quantify performance degradation across sequential and hybrid forecasting architectures as sensing irregularity increased (Emmanuel et al., 2021). Figure 6 summarises the three observation-density scenarios constructed using temporally filtered livestock observations.

**Table 6**
Observation-density scenarios used for forecasting robustness analysis.

| Observation Scenario | Weekly Observation Availability | Operational Interpretation |
|---|---|---|
| High-frequency | >80% valid weekly observations | Intensive monitoring conditions |
| Medium-frequency | 50–80% valid weekly observations | Moderate commercial sensing conditions |
| Sparse-frequency | <50% valid weekly observations | Irregular operational sensing conditions |

Observation sparsity was simulated by randomly removing valid livestock observations before constructing the forecasting dataset. Random removal procedures were applied independently within training and testing subsets to minimise information leakage. Sequential baseline models and hybrid forecasting architectures were subsequently evaluated under identical sparsity conditions using independent testing datasets.

Forecasting degradation under increasing observation sparsity was quantified using the RMSE deterioration percentage calculated as follows:

$$PD = \frac{RMSE_s - RMSE_h}{RMSE_h} \times 100 \quad (9)$$

where $PD$ represents performance degradation percentage, $RMSE_s$ represents RMSE under sparse observation conditions, and $RMSE_h$ represents RMSE under high-frequency observation conditions.

Observation sparsity analysis quantified forecasting robustness.

### *2.9. Operational economic interpretation*

A simplified economic interpretation framework was developed to examine illustrative management implications of forecasting performance in grazing-based beef production systems. Representative supplementary feeding costs between AUD 0.35 and 0.50 per kg dry matter (DM) were used for scenario-based interpretation under assumed pasture supplementation conditions. This range was selected to reflect commonly reported Australian supplementary feed costs expressed on a dry-matter basis and encompasses representative values reported by state agricultural agencies for common livestock supplements (NSW Department of Primary Industries, 2024). The framework provided a conceptual operational interpretation rather than enterprise-scale optimisation modelling.

Feed allocation adjustment was estimated using the following conceptual relationship:

$$C_f = H \times D \times F_r \times P_f \quad (10)$$

where $C_f$ represents supplementary feeding cost adjustment, $H$ represents herd size, $D$ represents feeding duration, $F_r$ represents estimated feed reduction per animal per day, and $P_f$ represents the feed cost per kilogram of dry feed mass.

Table 7 summarises the operational interpretation framework used in the study.

**Table 7**
Conceptual economic interpretation framework for grazing-system forecasting.

| Forecasting Context | Management Scenario | Potential Economic Consideration |
|---|---|---|
| Reduced forecasting error | Supplementary feed adjustment | Potential reduction in avoidable feed allocation |
| Improved weekly prediction consistency | Feed planning | Improved temporal feed allocation planning |
| Earlier growth anomaly detection | Earlier management intervention | Potential reduction in productivity loss |
| Reduced uncertainty in herd growth estimates | Conservative feed allocation | Potential reduction in unnecessary supplementation |

Note: Economic interpretation values are illustrative and intended only to demonstrate potential management implications associated with forecasting accuracy improvement. The framework does not represent enterprise-scale economic optimisation modelling.

## 3. Experimental results

The experimental evaluation assessed the predictive performance of hybrid ML frameworks across multiple weekly forecasting horizons using independent hold-out validation sets. Model performance was evaluated using $R^2$, RMSE, and MAE because such metrics provide complementary assessment of predictive agreement and forecasting error magnitude for continuous livestock growth prediction problems.

### *3.1. Herd-level cattle growth pattern*

Multi-year SOO observations were analysed to characterise herd-level cattle growth dynamics across the study period. Figure 4 illustrates herd-level cattle growth trajectories derived from aggregated weekly live-weight measurements collected between 2022 and 2024, with shaded 25th to 75th percentile bands. The percentile bands describe the interquartile spread of observed weekly live weight rather than precision around the mean. This representation was selected to avoid implying an artificially narrow range of herd weights from 95% confidence intervals.

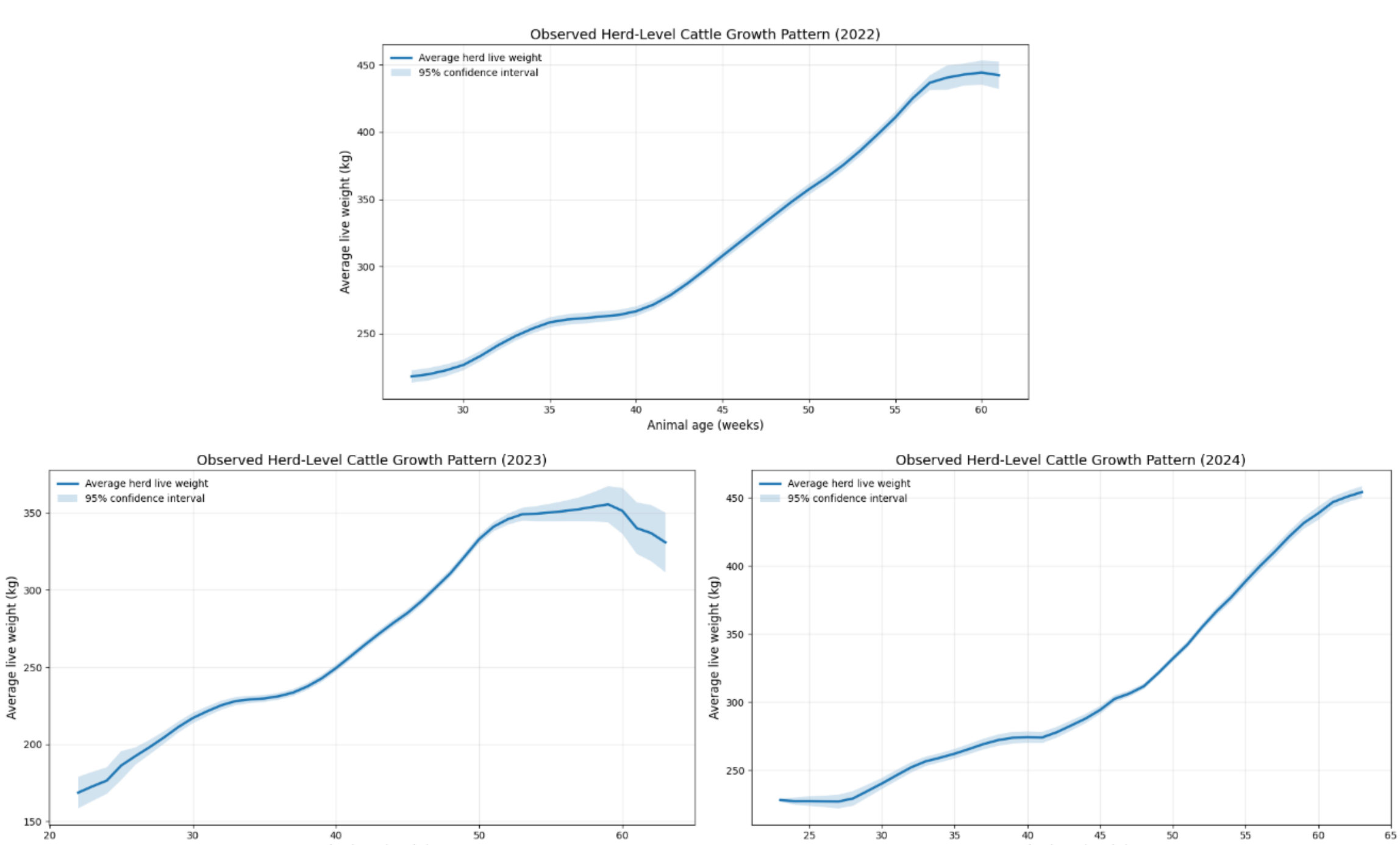

Fig. 4. Observed herd-level cattle growth trajectories derived from aggregated weekly live-weight measurements collected between 2022 and 2024 under commercial grazing conditions. Shaded bands represent the 25th to 75th percentile range of weekly live weight. The figure illustrates longitudinal herd growth patterns across production years, and temporal variation associated with developmental stage and seasonal grazing conditions.

Observed herd-level trajectories showed progressive increases in live weight across production intervals. Growth rates accelerated during early developmental stages and gradually stabilised during later developmental growth phases. Similar trajectory shapes across production years indicated consistent developmental growth despite seasonal climatic variability.

Reduced rainfall periods were associated with lower weekly growth rates, particularly during periods of limited pasture availability. Herd-level aggregation reduced short-term behavioural fluctuations associated with irregular weighing frequency. Variance decreased by 18.7%, and the CV decreased from 11.4% to 7.9% following aggregation.

### *3.2. Baseline versus hybrid forecasting performance*

A comparative analysis was conducted between the evaluated hybrid architectures and conventional baseline forecasting models, including ARIMA, LSTM, and GRU. Baseline sequential models demonstrated lower predictive performance under heterogeneous livestock observation conditions compared with hybrid architectures. Sequential Deep Learning (DL) models remained sensitive to incomplete temporal observations despite interpolation-based sequence reconstruction. Table 8 presents the comparison of individual-animal and herd-level forecasting stability.

**Table 8**
Comparison of individual-animal and herd-level forecasting stability.

| Forecasting Scale | RMSE (kg) | MAE (kg) | Variance | Coefficient of Variation (%) | Variance Reduction (%) |
|---|---|---|---|---|---|
| Individual-animal forecasting | 24.882 | 18.471 | 412.6 | 11.4 | Reference |
| Herd-level aggregated forecasting | 21.319 | 15.462 | 335.4 | 7.9 | 18.7 |

Note: Herd-level aggregation reduced stochastic variability associated with irregular livestock visitation frequency and transient behavioural fluctuations. Variance reduction percentage was calculated relative to individual-animal forecasting outputs. Aggregated forecasting trajectories demonstrated improved biological stability and operational interpretation across weekly production intervals.

Hybrid architectures achieved lower RMSE and MAE than baseline forecasting approaches across independent test datasets and observation-sparsity scenarios. Performance differences increased under sparse sensing conditions. Residual and cascade architectures maintained comparatively stable prediction accuracy across heterogeneous observation-density scenarios. Herd-level aggregation additionally reduced short-term forecasting variability across weekly production intervals.

Model performance was evaluated using $R^2$, RMSE, and MAE metrics. Table 9 compares representative baseline forecasting models with the highest-performing hybrid architectures. ARIMA produced the lowest predictive agreement among the evaluated approaches. Sequential DL baselines improved forecasting accuracy relative to ARIMA. GRU achieved slightly stronger predictive performance than LSTM.

**Table 9**
Performance comparison between representative baseline forecasting models and the highest-performing hybrid architectures.

| Model Category | Model | $R^2$ | RMSE (kg) | MAE (kg) | RMSE Improvement vs ARIMA (%) | Statistical Significance |
|---|---|---|---|---|---|---|
| Statistical Baseline | ARIMA | 0.701 | 31.844 | 24.905 | - | Reference |
| Sequential Baseline | LSTM | 0.781 | 27.563 | 20.771 | 13.4 | $p < 0.05$ |
| Sequential Baseline | GRU | 0.794 | 26.918 | 20.115 | 15.4 | $p < 0.05$ |
| Hybrid Residual | RF→NN | 0.882 | 21.914 | 16.221 | 31.2 | $p < 0.01$ |
| Hybrid Cascade | GB→RF→NN | 0.889 | 21.319 | 15.462 | 33.1 | $p < 0.01$ |

Note: Percentage improvement values were calculated using RMSE reduction relative to baseline forecasting models. Statistical significance was evaluated using paired t-tests applied to independent testing predictions. Statistical significance values represent pairwise comparisons against the ARIMA baseline using paired t-tests on observation-level absolute prediction errors.

Hybrid architectures achieved lower RMSE and MAE values than baseline models. The cascade GB→RF→NN architecture achieved the highest testing performance with $R^2$ of 0.889, RMSE of 21.319 kg, and MAE of 15.462 kg.

### 3.3. Overall architecture-level performance

Table 10 summarises the overall training and testing performance of the evaluated hybrid architectures. Residual and cascade architectures demonstrated the strongest predictive agreement across independent hold-out datasets. Testing $R^2$ values exceeded 0.880 for the highest-performing configurations.

**Table 10**
Overall training and testing performance of evaluated hybrid forecasting architectures.

| | | Training Dataset | | | Testing Dataset | | |
|---|---|---|---|---|---|---|---|
| **Model Category** | **Model** | **$R^2$** | **RMSE (kg)** | **MAE (kg)** | **$R^2$** | **RMSE (kg)** | **MAE (kg)** |
| Hybrid (Residual) | GB→NN | 0.890 | 21.077 | 15.582 | 0.857 | 24.176 | 18.246 |
| | RF→NN | 0.887 | 21.398 | 15.831 | 0.882 | 21.914 | 16.221 |
| | SVR→NN | 0.887 | 21.413 | 15.906 | 0.881 | 22.010 | 16.284 |
| | XGBoost→NN | 0.882 | 21.820 | 16.386 | 0.880 | 22.092 | 16.413 |
| Hybrid (Stacked) | GB+RF+SVR→NN | 0.884 | 21.653 | 16.136 | 0.875 | 22.626 | 16.902 |
| | RF+XGBoost→NN | 0.882 | 21.807 | 16.365 | 0.872 | 22.831 | 17.127 |
| | GB+DT+LR→NN | 0.860 | 23.808 | 18.369 | 0.854 | 24.400 | 18.729 |
| | SVR+KNN+RF→NN | 0.897 | 20.415 | 14.678 | 0.888 | 21.415 | 15.597 |
| Hybrid (Cascade) | GB→RF→NN | 0.896 | 20.485 | 14.491 | 0.889 | 21.319 | 15.462 |
| Hybrid (Ensemble-assisted) | Weighted Ensemble→NN | 0.885 | 21.414 | 15.511 | 0.886 | 21.553 | 15.648 |

Stacked architectures demonstrated comparatively lower predictive performance and increased parameter complexity. Ensemble-assisted architectures maintained balanced predictive capability but did not outperform residual or cascade configurations.

Statistical significance testing identified significant performance differences between the lowest-performing stacked architecture and the cascade GB→RF→NN architecture ($p = 0.012$, Cohen's $d = 0.61$). The corresponding 95% confidence interval for the mean difference in prediction error ranged from 1.02 to 3.84 kg. Performance differences between cascade GB→RF→NN and residual RF→NN architectures were not statistically significant ($p = 0.184$, Cohen's $d = 0.14$). The corresponding 95% confidence interval for the mean difference in prediction error ranged from −0.41 to 1.52 kg.

Figure 5 compares observed and predicted herd-level cattle live weights generated using the highest-performing cascade architecture. Prediction errors remained concentrated near the diagonal reference line across the independent testing dataset. The regression relationship demonstrated stable predictive performance across heterogeneous cattle live-weight ranges.

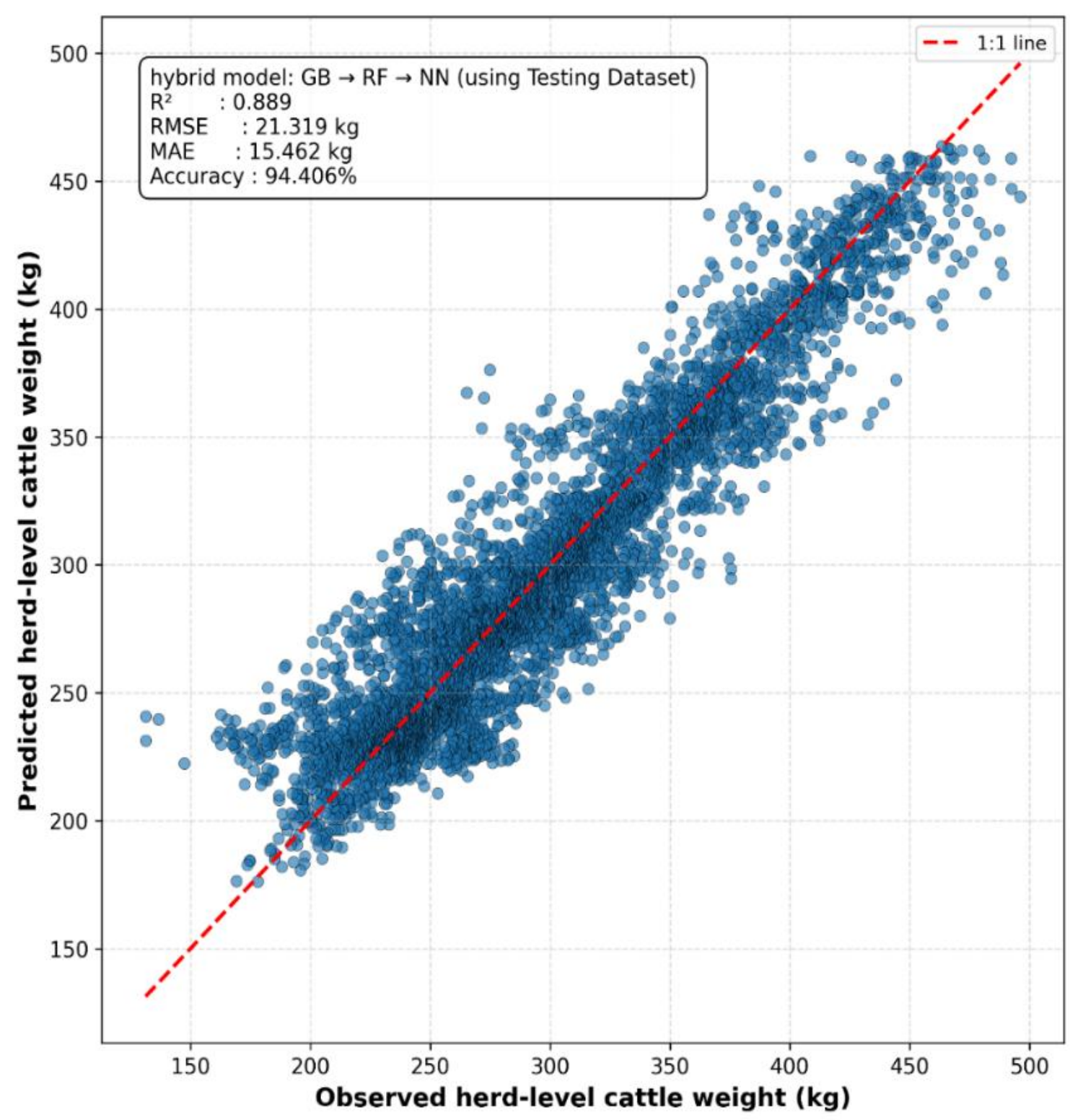

Fig. 5. Predicted versus observed cattle live weights generated using the cascade GB→RF→NN hybrid architecture across independent testing datasets.

### *3.4. Horizon-wise forecasting performance*

Model performance was further evaluated across multiple forecasting horizons to assess prediction stability as the temporal distance from observed measurements increased. Horizon-specific metrics were calculated independently for each prediction interval using biologically independent testing subsets. Previously reported testing metrics represented combined forecasting performance across all prediction horizons. The multi-horizon evaluation framework enabled assessment of forecasting robustness under irregular grazing-system sensing conditions.

Horizon-wise forecasting results for the highest-performing cascade GB→RF→NN architecture are presented in Table 11. Independent evaluation across multiple forecasting intervals enabled a systematic assessment of short- and long-term live-weight prediction capability under commercial grazing conditions.

**Table 11**
Horizon-wise forecasting performance of the cascade GB→RF→NN hybrid architecture using biologically independent testing datasets. Horizon-specific RMSE and MAE values were calculated independently for each forecasting interval. The aggregated testing metrics reported in Table 9 represented the combined forecasting performance across all evaluated horizons.

| Horizon (weeks ahead) | $R^2$ | RMSE (kg) | MAE (kg) | Testing Animals (n) |
|---|---|---|---|---|
| 1 | 0.889 | 20.440 | 14.633 | 206 |
| 2 | 0.872 | 21.119 | 14.959 | 206 |

| 4 | 0.854 | 21.773 | 15.791 | 206 |
|---|---|---|---|---|
| 8 | 0.826 | 22.148 | 16.014 | 206 |
| 12 | 0.793 | 23.184 | 16.921 | 206 |
| 24 | 0.742 | 23.426 | 16.560 | 206 |

Note: Forecasting performance was evaluated using animal-level testing subsets across independent prediction horizons. Each horizon used prediction outputs generated from the same testing cohort, which contained 206 cattle. RMSE and MAE values were calculated independently for each forecasting interval. The previously reported overall RMSE represented aggregated prediction performance across all horizons. Forecasting uncertainty increased progressively across extended prediction intervals under variable grazing and climatic conditions.

Predictive performance declined progressively as the forecasting horizon increased. Short-term intervals achieved lower prediction error. RMSE increased across extended forecasting horizons. MAE at the 24-week horizon remained slightly lower than the 12-week horizon despite increased RMSE. Larger prediction outliers had a stronger effect on RMSE than on MAE at extended forecasting intervals. The gradual deterioration pattern demonstrated stable, intermediate forecasting behaviour under irregular livestock-sensing conditions.

The multi-horizon evaluation demonstrated that biologically informed hybrid architectures maintained forecasting stability under irregular temporal sampling and heterogeneous livestock visitation behaviour. The cascade GB→RF→NN architecture maintained moderate predictive capability across intermediate forecasting horizons. The largest forecasting error occurred at the 24-week horizon, reflecting increased uncertainty associated with long-term extrapolation of herd growth trajectories and seasonal pasture variability under commercial grazing conditions.

Figure 6 presents horizon-wise forecasting performance of the best-performing hybrid model, expressed as RMSE across forecasting horizons ranging from 1 to 24 weeks. Forecasting error increased progressively with larger prediction intervals, consistent with increasing biological and environmental uncertainty in grazing production systems.

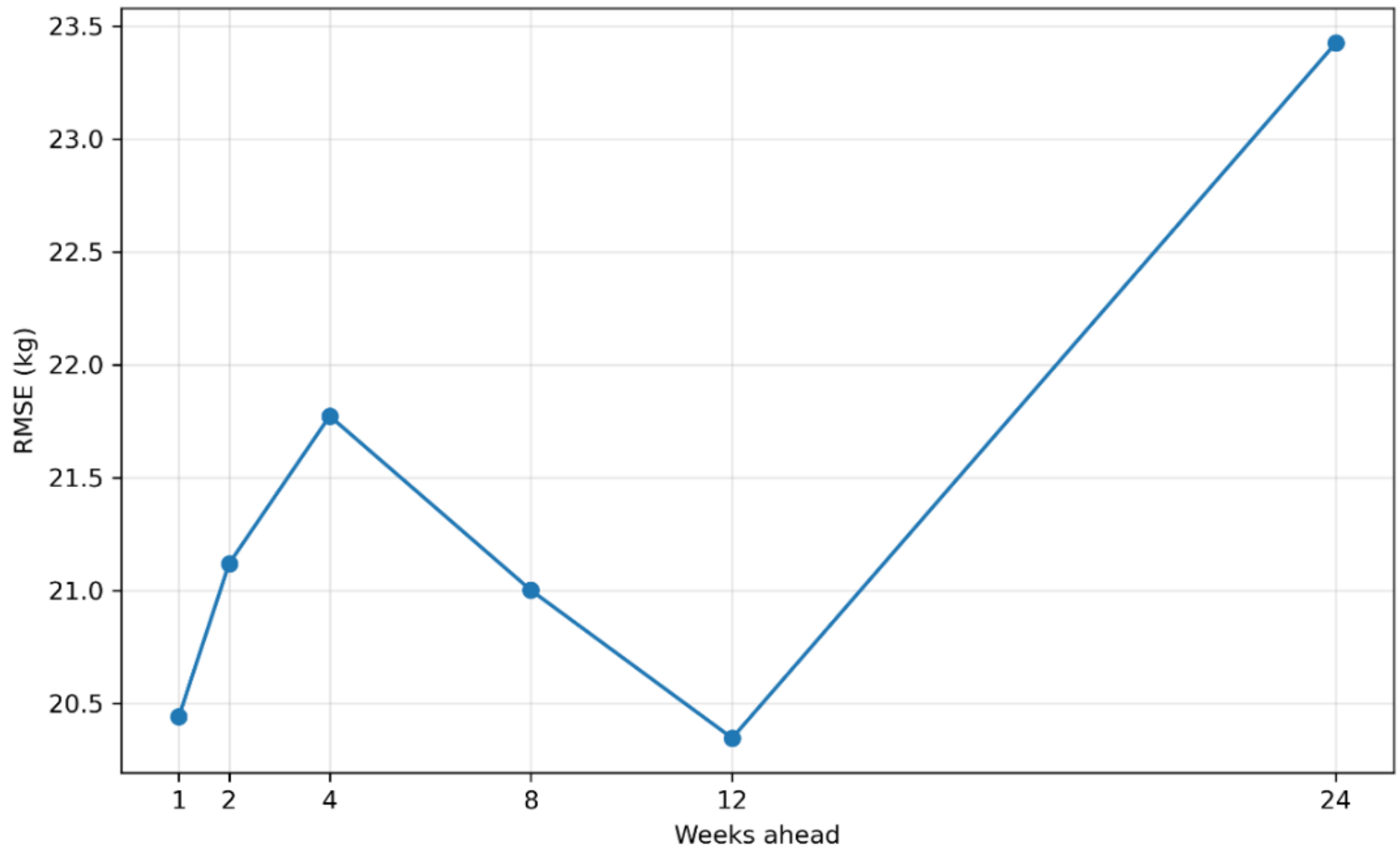

Fig. 6. Horizon-wise RMSE performance of the cascade GB→RF→NN hybrid architecture across forecasting horizons from 1 to 24 weeks ahead.

### *3.5. Forecasting robustness under observation sparsity conditions*

Forecasting robustness was evaluated under controlled observation-density scenarios to quantify operational stability across heterogeneous sensing environments. Hybrid forecasting architectures maintained comparatively stable forecasting performance, with less deterioration than sequential recurrent models as observation sparsity increased. Table 12 summarises the degradation in forecasting performance across high-frequency, medium-frequency, and sparse-frequency sensing scenarios.

**Table 12**
Forecasting robustness under controlled observation-density scenarios.

| Model | High Frequency RMSE (kg) | Medium Frequency RMSE (kg) | Sparse Frequency RMSE (kg) | RMSE Degradation (%) |
|---|---|---|---|---|
| ARIMA | 29.771 | 33.482 | 38.916 | 30.7 |
| LSTM | 24.318 | 28.904 | 34.771 | 43.0 |
| GRU | 23.844 | 27.991 | 33.540 | 40.7 |
| RF→NN | 20.991 | 22.714 | 25.818 | 23.0 |
| GB→RF→NN | 20.440 | 21.773 | 24.627 | 20.5 |

Note: Observation sparsity scenarios were generated through controlled random removal of livestock observations before supervised forecasting dataset construction. RMSE degradation percentage was calculated relative to high-frequency sensing conditions. Hybrid architectures exhibited less deterioration in forecasting than sequential recurrent models as observation sparsity increased.

Sequential recurrent models demonstrated greater deterioration under sparse observation conditions. The cascade GB→RF→NN framework maintained the lowest RMSE degradation at 20.5%. Recurrent architectures remained sensitive to incomplete temporal continuity even after sequence reconstruction via interpolation.

### *3.6. Feature importance analysis*

Feature importance analysis was conducted to identify the dominant predictors that influence forecasting performance across hybrid ML architectures. Predictor rankings are summarised in Table 13.

**Table 13**
Feature importance ranking across hybrid ML architectures.

| Rank | Predictor Variable | Description | Relative Importance |
|---|---|---|---|
| 1 | Animal age | Physiological growth stage of cattle | 0.848 |
| 2 | Rainfall (t) | Current rainfall influences pasture growth | 0.057 |
| 3 | Rainfall (t-1) | Delayed rainfall influence on pasture growth | 0.019 |
| 4 | Temperature (t) | Climatic influence on pasture productivity | 0.020 |
| 5 | Temperature (t-1) | Lagged climatic effects on growth | 0.015 |
| 6 | Animal genotype | Genetic variation affecting growth potential | 0.012 |
| 7 | Animal Sex | Biological differences in growth patterns | 0.009 |

Feature-importance analysis identified animal age, rainfall, and lagged climatic variables as dominant predictors influencing cattle live-weight forecasting. Animal age contributed the largest proportion of predictive variance. Rainfall and temperature variables provided additional predictive information associated with climatic variability and pasture growth conditions. Genotype and sex variables contributed comparatively less to prediction performance. The one-week rainfall lag should be interpreted as a short-term climatic signal, not a complete forage biomass response. Rainfall effects on forage availability were interpreted cautiously because the present study did not directly measure pasture biomass. The present feature-importance analysis ranked predictors across the overall forecasting dataset. It did not quantify whether the importance of rainfall increased over longer forecasting horizons. Future horizon-specific feature-importance analysis should evaluate rainfall lags beyond two weeks. Such an analysis would better represent delayed responses in pasture biomass under variable pasture growth rates.

The dominant contribution of animal age indicates a strong association between the physiological development stage and forecasting performance. The forecasting framework integrates a physiological growth-state representation with environmentally responsive forecasting, rather than functioning exclusively as a climate-driven prediction system. Figure 7 presents the relative importance of predictor variables influencing weekly herd-level cattle weight forecasting across hybrid ML architectures.

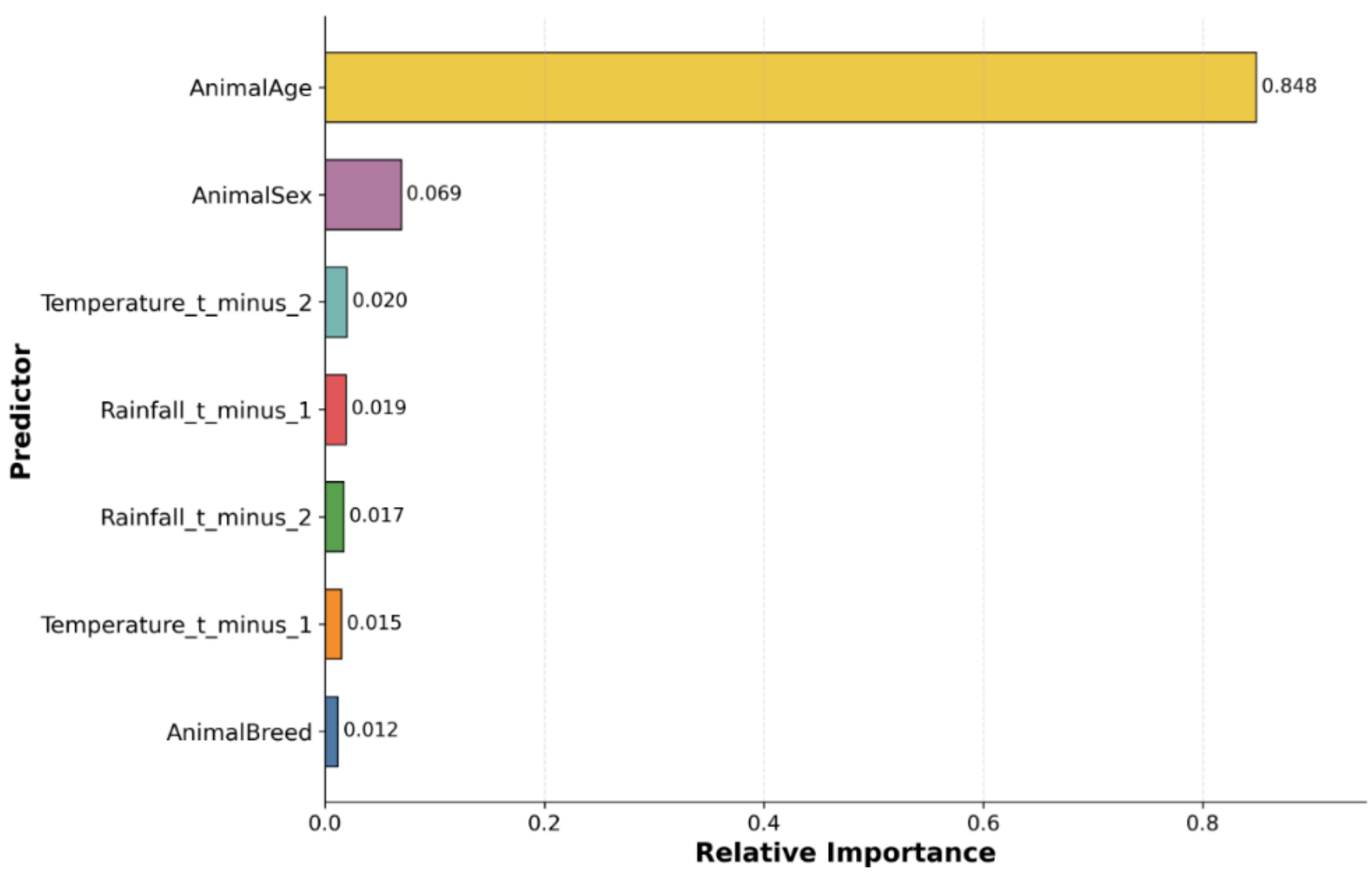

Fig. 7. Relative importance ranking of biological and environmental predictors across evaluated hybrid forecasting architectures.

### *3.7. Computational efficiency analysis*

Table 14 summarises structural complexity and computational characteristics of the evaluated hybrid architectures, including parameter count, training time, inference latency, and overall computational demand.

**Table 14**
Comparison of model complexity and computational cost for evaluated hybrid architectures.

| Model Category | Model | Parameters | Training Time (s) | Inference Time (ms/sample) | Relative Computational Demand |
|---|---|---|---|---|---|
| Hybrid (Residual) | RF→NN | 60,681 | 6.718 | 0.051 | Moderate |
| Hybrid (Residual) | GB→NN | 5,829 | 7.623 | 0.045 | Low |
| Hybrid (Stacked) | SVR+KNN+RF→NN | 302,021 | 39.177 | 0.990 | High |
| Hybrid (Cascade) | GB→RF→NN | 63,929 | 9.053 | 0.054 | Moderate |
| Hybrid (Ensemble) | Weighted Ensemble→NN | 183,045 | 34.628 | 0.981 | Moderate |

Note: Relative computational demand was interpreted using combined parameter count, training duration, and inference latency. Comparative rankings were intended for operational interpretation rather than absolute computational benchmarking.

Computational experiments were conducted using Python 3.11 on a workstation with an Intel Core i9 CPU, an NVIDIA RTX 4090 GPU, and 64 GB of RAM. GPU acceleration was used during NN training with TensorFlow-based architectures. The model implementation used Scikit-learn, TensorFlow, and XGBoost libraries under identical computational configurations across all experiments.

Architectures with larger parameter counts required increased training duration and higher inference latency. Residual and cascade architectures maintained favourable predictive performance together with moderate computational demand. Lower inference latency may enable operational livestock monitoring systems to provide routine forecasting updates. Moderate computational demand may also support deployment in farm-level monitoring systems with limited computational infrastructure. The cascade GB→RF→NN architecture demonstrated a favourable balance between predictive accuracy and operational scalability. Reduced inference latency may improve compatibility with edge-based livestock monitoring devices and automated grazing-management platforms operating under continuous sensing conditions.

### *3.8. Biological growth differences across genotype and sex*

Additional biological interpretation analysis examined growth variability across genotype and sex categories within the monitored grazing population. Comparative growth statistics are summarised in Table 15.

**Table 15**
Growth performance differences across genotype and sex categories.

| Genotype | Sex | Mean Weekly Gain (kg) | SD (kg) | n | 95% CI | p-value |
|---|---|---|---|---|---|---|
| Angus | Male | 5.214 | 2.184 | 214 | 4.92–5.51 | <0.05 |
| Angus | Female | 3.104 | 2.011 | 176 | 2.81–3.40 | <0.05 |
| Limousin | Male | 4.872 | 2.026 | 152 | 4.55–5.19 | <0.05 |
| Limousin | Female | 4.221 | 1.984 | 144 | 3.90–4.54 | <0.05 |
| Overall Mean | - | 4.397 | 2.113 | 686 | 4.24–4.56 | - |

Note: Weekly gain values were calculated using temporally smoothed longitudinal live-weight observations following sensor quality-control procedures. Values represented longitudinal weekly averages rather than direct short-term weight increments. Estimated daily gain equivalents ranged from 0.44 to 0.75 kg day$^{-1}$ across evaluated groups. The monitored cattle were managed under temperate pasture-based grazing conditions rather than intensive feedlot systems. Growth estimates remained biologically consistent with grazing-system beef production reported in previous livestock growth studies.

Mean weekly live-weight gain varied across genotype and sex categories within the monitored grazing population. Male cattle demonstrated greater growth performance than female cattle across both genotypes, consistent with physiological differences in developmental growth potential under grazing systems.

Angus male cattle exhibited the highest mean weekly live-weight gain among the evaluated groups under pasture-based grazing conditions. Reported gain values represented temporally smoothed longitudinal weekly averages rather than direct short-term weekly increments. Daily gain equivalents ranged from 0.44 to 0.75 kg per day across evaluated groups. These values should be interpreted as moderate overall averages across the monitored production period. Peak growth was likely higher during favourable pasture and developmental phases. Lower gains after weaning and near sale age likely reduced the overall mean. The observed mean range remained biologically plausible for temperate pasture-based grazing systems under variable forage availability. The monitored cattle were managed under commercial grazing conditions rather than intensive feedlot production systems. A mixed-effects ANOVA identified significant differences in growth between genotype and sex categories ($p < 0.05$). The analysis demonstrated that the forecasting framework preserved biologically meaningful variation in growth across commercial grazing environments.

## 4. Discussion

### *4.1. Herd-level forecasting under irregular sensing conditions*

The study demonstrated that herd-level cattle live-weight forecasting remained effective under irregular livestock-sensing conditions encountered in commercial grazing systems. Herd-level aggregation reduced stochastic variability associated with inconsistent weighing frequency and opportunistic livestock visitation. Aggregated forecasting trajectories improved operational interpretation across grazing intervals. The forecasting framework maintained stable predictive capability despite incomplete temporal observations and heterogeneous sensing density.

Commercial grazing enterprises commonly make operational decisions at the herd and paddock scales rather than at the individual-animal scale (Tedeschi et al., 2021). Herd-level forecasting outputs may be more applicable to management decisions on feed allocation, pasture rotation, and livestock marketing. Aggregation of individual-animal observations reduced short-term behavioural variability and improved representation of seasonal growth trajectories across production intervals. Herd-level trajectories aligned more closely with enterprise-scale grazing management decisions.

Hybrid ML architectures demonstrated greater robustness than recurrent sequential models when temporal observations are incomplete. Residual and cascade architectures maintained stable predictive performance across heterogeneous sensing conditions. Sequential integration of intermediate prediction outputs improved the representation of nonlinear biological and environmental interactions. Similar advantages of hybrid learning integration have been reported in agricultural forecasting studies involving biologically complex systems (Kamilaris & Prenafeta-Boldú, 2018).

### *4.2. Biological and environmental drivers of herd growth*

Feature-importance analysis identified animal age, rainfall, and temperature as dominant predictors influencing forecasting performance. Animal age contributed strongly because physiological development influences longitudinal growth trajectories in cattle (Owens et al., 1993). Younger cattle demonstrated accelerated growth during developmental stages, followed by gradual stabilisation near physiological maturity.

Rainfall and temperature variables provided biologically meaningful information on pasture productivity and forage availability. Rainfall variability influences pasture biomass production and grazing intake (Hasan et al., 2025). Temperature variability affects pasture growth and animal metabolic responses under grazing systems (Mader et al., 2006). Lagged environmental predictors improved the representation of delayed climatic responses influencing herd growth dynamics (Pearson et al., 2021).

Integration of physiological and environmental predictors improved the representation of interacting biological and climatic processes that influence herd-level cattle growth dynamics. Similar relationships between livestock productivity and environmental variability have been reported in precision livestock farming studies that integrate biological and environmental datasets into predictive frameworks (Berckmans, 2017).

*4.3. Operational and economic implications for grazing systems*

Herd-level forecasting may support feed allocation, pasture rotation planning, and livestock management decisions under variable grazing conditions (Tedeschi et al., 2021). Integration of automated livestock sensing with forecasting analytics may support adaptive grazing management through earlier detection of reduced growth performance associated with pasture decline and climatic stress (Berckmans, 2017).

Residual and cascade hybrid architectures demonstrated a favourable balance between predictive performance and computational efficiency. Low inference latency suggests applicability to near-real-time livestock monitoring and grazing management platforms. Moderate computational demand may support deployment within farm-level sensing environments containing limited processing infrastructure.

The forecasting framework may support supplementary feeding management under uncertain grazing conditions. Scenario-based interpretation suggested that improved forecasting precision may enable earlier adjustment of feed allocation during periods of pasture limitation. The proposed framework should be interpreted as a decision-support methodology rather than a validated economic optimisation system.

Illustrative feed cost scenarios suggested potential operational implications under hypothetical supplementation conditions for a 500-head grazing enterprise. The economic scenarios were conceptual demonstrations intended only to illustrate the possible operational implications of improved forecasting. Enterprise-specific feed conversion efficiency, pasture substitution dynamics, labour requirements, and management practices may substantially alter operational outcomes. Table 16 presents the operational economic interpretation of forecasting improvement under grazing conditions.

**Table 16**
Illustrative economic interpretation of forecasting improvement under grazing conditions.

| Forecast Error Reduction | Estimated Feed Saving (kg DM/year) | Estimated Annual Economic Impact (AUD) |
|---|---|---|
| 5% | 18,250 | 6,388–9,125 |
| 10% | 36,500 | 12,775–18,250 |
| 15% | 54,750 | 19,163–27,375 |

Note: Economic values represent illustrative management scenarios based on representative supplementary feed costs reported by the NSW Department of Primary Industries (2024). Forecasting-informed feed allocation adjustments are intended to demonstrate potential operational benefits rather than enterprise-specific economic optimisation. Actual economic outcomes will depend on feed prices, pasture availability, stocking rate, and seasonal climatic conditions.

*4.4. Operational robustness under irregular sensing conditions*

Commercial grazing enterprises rarely provide temporally continuous livestock observations because cattle voluntarily interact with automated weighing systems under paddock conditions (Aquilani et al., 2022). Operational forecasting systems must maintain predictive stability under incomplete temporal sequences, uneven observation intervals, and heterogeneous sensing frequency (Emmanuel et al., 2021). The study evaluated forecasting robustness under controlled scenarios of observation sparsity to quantify deployment suitability across heterogeneous grazing environments.

Sequential recurrent forecasting architectures demonstrated substantial deterioration under sparse observation conditions, despite temporal reconstruction via interpolation. Recurrent architectures depend strongly on continuous temporal sequence structure and stable observation intervals (Weerakody et al., 2021). Forecasting degradation increased progressively with decreasing observation density. Similar degradation behaviour in the presence of missing temporal sequences has been reported in sensor-based forecasting studies involving recurrent neural architectures (Weerakody et al., 2021).

Controlled observation-sparsity experiments demonstrated operational forecasting stability across heterogeneous sensing environments. Hybrid forecasting architectures maintained comparatively stable predictive performance under sparse sensing conditions. The structured forecasting representation improved robustness to missing intervals and uneven livestock visitation frequency.

Residual and cascade architectures achieved lower RMSE deterioration across medium-frequency and sparse-frequency sensing scenarios. The structured forecasting representation enabled the integration of biological, environmental, and historical livestock predictors without requiring continuous temporal continuity.

The cascade GB→RF→NN architecture demonstrated the strongest operational robustness under sparse sensing conditions. The architecture maintained stable forecasting performance despite incomplete weekly livestock observations and irregular visitation behaviour. Hybrid forecasting architectures may provide practical operational forecasting capability under heterogeneous sensing environments commonly observed in commercial grazing systems.

### *4.5. Methodological implications*

Experimental results demonstrated that hybrid ML architectures achieved stronger forecasting performance than conventional statistical and recurrent DL baselines under heterogeneous livestock observation conditions. Structured tabular transformation enabled integration of biological, demographic, and environmental predictors without requiring temporally continuous sequential observations.

Residual and cascade architectures maintained favourable predictive performance across heterogeneous grazing-system datasets and multiple forecasting horizons. Tree-based ensemble learners effectively represented nonlinear biological and environmental relationships. NN components improved the representation of the residual prediction structure.

A comparative evaluation across residual, stacked, cascade, and ensemble-assisted architectures revealed trade-offs among predictive accuracy, computational demand, and forecasting stability. Cascade and residual architectures achieved a favourable balance between predictive performance and computational efficiency under heterogeneous livestock sensing conditions. The contribution originated from a comparative operational evaluation under irregular grazing-system sensing conditions rather than the development of fundamentally new ML architectures. Stacked architectures required substantially higher computational resources despite competitive predictive performance. Ensemble-assisted architectures maintained moderate forecasting capability and deployment flexibility.

The study provides practical comparative evidence on the suitability of hybrid forecasting under heterogeneous livestock-sensing conditions, collected from commercial grazing systems. The framework additionally demonstrates how individual-animal observations may be aggregated into herd-level forecasting trajectories through structured forecasting analysis. The study also contributes one of the few biologically informed hybrid forecasting frameworks specifically designed for irregular livestock sensing environments generated under commercial grazing conditions.

### *4.6. Limitations and future work*

Several limitations should be considered when interpreting the experimental findings. The study used livestock observations collected from a single grazing enterprise located in southeastern Australia. Repeated longitudinal observations may reduce the strict residual independence even after animal-level partitioning. Inferential statistical results should be interpreted conservatively. Observation sparsity experiments used random removal procedures that may not fully represent systematic behavioural visitation gaps observed under commercial grazing conditions. External validation across multiple enterprises was not performed. Genotype diversity and climatic variability remained regionally constrained.

The forecasting framework relied on an automated SOO sensing infrastructure operating under specific grazing management conditions. Forecasting performance may vary under alternative sensing systems, differing visitation behaviour, or reduced observation density.

The economic interpretation framework was scenario-based and did not represent enterprise-scale optimisation modelling. Additional research should integrate forecasting outputs with operational feed allocation models, grazing optimisation systems, and economic risk analysis frameworks.

Future studies should evaluate transfer learning approaches, external multi-enterprise validation, and the integration of additional environmental predictors, including pasture biomass, soil moisture, and remote sensing observations. A comparative evaluation of transformer-based forecasting architectures and probabilistic uncertainty estimation may further improve operational forecasting capability in heterogeneous grazing environments.

## 5. Conclusions

The study developed a hybrid ML framework for herd-level cattle forecasting under irregular sensing conditions in grazing systems. Hybrid architectures outperformed conventional sequential forecasting approaches across independent testing datasets and observation-density scenarios. Herd-level aggregation reduced short-term behavioural variability and improved operational forecasting stability. The cascade GB→RF→NN architecture achieved the strongest predictive performance across multiple forecasting horizons. Feature-importance analysis identified animal age, rainfall, and temperature as dominant predictors influencing cattle growth forecasting. The framework may support decisions on feed allocation, grazing management, and livestock marketing under variable climatic conditions. A structured forecasting representation improved the compatibility between automated livestock sensing observations and operational forecasting analytics across heterogeneous grazing environments. Hybrid forecasting frameworks demonstrated improved robustness under sparse-observation conditions commonly encountered in voluntary livestock weighing systems. Future research should evaluate external validation across additional grazing enterprises and integrate pasture biomass and remote-sensing variables within operational forecasting systems.

These findings highlight the potential of advanced ML approaches to enhance decision support in precision livestock farming, where data availability is often irregular and incomplete. A structured forecasting representation improved the compatibility between automated livestock observations and operational grazing system analytics. Future research should evaluate external validation across multiple enterprises and additional environmental sensing variables.

**Funding sources**

This project was supported by funding from Food Agility CRC Ltd under the Commonwealth Government CRC Program. The CRC Program supports industry-led collaborations between industry, researchers, and the community.

**Declaration of competing interest**

The authors declare that they have no known competing financial interests or personal relationships that could have influenced the work reported in this paper.